# Mutual information for the selection of relevant variables in spectrometric nonlinear modelling

F. Rossi[1], A. Lendasse[2], D. François[3], V. Wertz[3], M. Verleysen[4*]

[1] Projet AxIS, INRIA-Rocquencourt, Domaine de Voluceau, Rocquencourt, B.P. 105, F-78153 Le Chesnay Cedex, France, Fabrice.Rossi@inria.fr.
[2] Helsinki University of Technology – Lab. Computer and Information Science, Neural Networks Research Centre, P.O. Box 5400, FIN-02015 HUT, Finland, lendasse@hut.fi
[3] Université catholique de Louvain – Machine Learning Group, CESAME, 4 av. G. Lemaître, B-1348 Louvain-la-Neuve, Belgium, francois@auto.ucl.ac.be
[4] Université catholique de Louvain – Machine Learning Group, DICE, 3 place du Levant, B-1348 Louvain-la-Neuve, Belgium, verleysen@dice.ucl.ac.be and Université Paris I Panthéon Sorbonne, SAMOS-MATISSE, 90 rue de Tolbiac F-75634 Paris Cedex 13, France

**Cooresponding author**

Michel Verleysen

Email : verleysen@dice.ucl.ac.be

Place du Levant 3,
B-1348 Louvain-La-Neuve
Belgium

[*] MV is a Senior Research Associate of the Belgian F.N.R.S. (National Fund For Scientific Research). The work of DF is funded by a grant from the Belgian FRIA. Part of the work of DF and the work of VW is supported by the Interuniversity Attraction Pole (IAP), initiated by the Belgian Federal State, Ministry of Sciences, Technologies and Culture. Part the work of A. Lendasse is supported by the project New Information Processing Principles, 44886, of the Academy of Finland. The scientific responsibility rests with the authors.

**Abstract**

Data from spectrophotometers form vectors of a large number of exploitable variables. Building quantitative models using these variables most often requires using a smaller set of variables than the initial one. Indeed, a too large number of input variables to a model results in a too large number of parameters, leading to overfitting and poor generalization abilities. In this paper, we suggest the use of the mutual information measure to select variables from the initial set. The mutual information measures the information content in input variables with respect to the model output, without making any assumption on the model that will be used; it is thus suitable for nonlinear modelling. In addition, it leads to the selection of variables among the initial set, and not to linear or nonlinear combinations of them. Without decreasing the model performances compared to other variable projection methods, it allows therefore a greater interpretability of the results.

# 1. Introduction

Many analytical problems related to spectrometry require predicting a quantitative variable through a set of measured spectral data. For example, one can try to predict a chemical component concentration in a product through its measured infrared spectrum. In recent years, the importance of such problems in various fields including the pharmaceutical [1-2], food [3] and textile industries [4] have grown dramatically. The chemical analysis by spectrophotometry rests on the fast acquisition of a great number of spectral data (several hundred, even several thousands).

A remarkable characteristic of many such high-dimensional data is the so-called *colinearity*. This means that some of the spectral variables may be quite perfect linear combinations of other ones. In addition, it is known that a model cannot include more effective parameters than there are available samples to learn the model; otherwise, *overfitting* may appear, meaning that the model can behave efficiently on the learning data but poorly on other ones, making it useless. In spectrometric problems, one is often faced with databases having more variables (spectra components) than samples; and almost all models use at least as many parameters as the number of input variables. These two problems, colinearity and risk of overfitting, already exist in linear models. However, their effect may be even more dramatic when nonlinear models are used (there are usually more parameters than in linear models, and the risk of overfitting is higher).

In such high-dimensional problems, it is thus necessary to use a smaller set of variables than the initial one. There are two ways to achieve this goal. The first one is to *select* some variables among the initial set; the second one is to *replace* the initial set by another, smaller, one, with new variables that are linear or nonlinear combinations of the initial ones. The second method is more general than the first one; it allows more flexibility in the choice of the new variables, which is a potential advantage. However, there are two drawbacks associated to this flexibility. The first one is the fact that projections (combinations) are easy to define in a linear context, but are much more difficult to design in a nonlinear one [5]. When using nonlinear models built on the new variables, it is obvious that linearly built sets of variables are not appropriate. The second one resides in the interpretability; while initial spectral variables (defined at known wavelengths) are possible to interpret from a chemometrics point of view, linear, or even worst nonlinear, combinations of them are much more difficult to handle.

Despite its inherent lack of generality, the selection of spectral data among the initial ones may thus be preferred to projection methods. When used in conjunction with nonlinear models, it also provides interpretability which usually is difficult to obtain with nonlinear models, often considered as "black-boxes".

This paper describes a method to select spectral variables by using a concept form information theory: the measure of *mutual information*. Basically, the mutual information measures the amount of information contained in a variable or a group of variables, in order to predict the dependent one. It has the unique advantage to be model-independent and nonlinear at the same time. Model-independent means that no assumption is made about the model that will be used on the spectral variables selected by mutual information; nonlinear means that the mutual information measures the nonlinear relationships between variables, contrarily to the correlation that only measures the linear relations.

In a previous paper [6] the possibility to select the first variable of the new selected set using the mutual information concept was described. Nevertheless, due to the lack of reliable methods to estimate the mutual information for groups of variables, the next variables from the second one were chosen by a forward-backward procedure. The latter, which is model-dependent, is extremely heavy from a computational point of view, making it unpractical when nonlinear models are used. In this paper, we extend the method suggested in [6] to the selection of all variables by using an estimator of the mutual information recently proposed in the literature [7]. The method is illustrated on two near-infrared spectroscopy benchmarks.

In the following of this paper, the concept of mutual information is described, together with methods to estimate it (Section 2). Then the use of this concept to order and select variables in a spectrometry model is described in Section 3. Section 4 presents the nonlinear models that have been used in the experiments (Radial-Basis Function Networks and Least-Square Support Vector Machines), together with statistical resampling methods used to choose their complexity and to evaluate their performances in an objective way. Finally, Section 5 presents the experimental results on two near-infrared spectroscopy benchmarks.

## 2. Mutual information

### 2.1 Definitions

The first goal of a prediction model is to minimize the uncertainty on the dependent variable. A good formalization of the uncertainty of a random variable is given by Shannon and Weaver's [8] information theory. While first developed for binary variables, it has been extended to continuous variables. Let $X$ and $Y$ be two random variables (they can have real or vector values). We denote $\mu_{X,Y}$ the joint probability density function (pdf) of $X$ and $Y$. We recall that the marginal density functions are given by

$$\mu_X(x) = \int \mu_{X,Y}(x,y) dy \tag{1}$$

and

$$\mu_Y(y) = \int \mu_{X,Y}(x,y) dx . \tag{2}$$

Let us now recall some elements of information theory. The uncertainty on $Y$ is given by its entropy defined as

$$H(Y) = -\int \mu_Y(y) \log \mu_Y(y) dy . \tag{3}$$

If we get knowledge on $Y$ indirectly by knowing $X$, the resulting uncertainty on $Y$ knowing $X$ is given by its conditional entropy, that is

$$H(Y|X) = -\int \mu_X(x) \int \mu_Y(y|X=x) \log \mu_Y(y|X=x) dy dx . \tag{4}$$

The joint uncertainty of the (X,Y) pair is given by the joint entropy, defined as

$$H(X,Y) = -\int \mu_{X,Y}(x,y) \log \mu_{X,Y}(x,y) dx dy . \tag{5}$$

The mutual information between $X$ and $Y$ can be considered as a measure of the amount of knowledge on $Y$ provided by $X$ (or conversely on the amount of knowledge on $X$ provided by $Y$). Therefore, it can be defined as [9]:

$$I(X,Y) = H(Y) - H(Y|X) , \tag{6}$$

which is exactly the reduction of the uncertainty of Y when X is known. If Y is the dependant variable in a prediction context, the mutual information is thus particularly suited to measure the pertinence of X in a model for Y [10]. Using the properties of the entropy, the mutual information can be rewritten into

$$I(X,Y) = H(X) + H(Y) - H(X,Y), \tag{7}$$

that is, according to the previously recalled definitions, into [11]:

$$I(X,Y) = \int \mu_{X,Y}(x,y) \log \frac{\mu_{X,Y}(x,y)}{\mu_X(x)\mu_Y(y)} dxdy. \tag{8}$$

Therefore we only need to estimate $\mu_{X,Y}$ in order to estimate the mutual information between X and Y by (1), (2) and (8).

## 2.2. Estimation of the mutual information

As detailed in the previous section, estimating the mutual information (MI) between X and Y requires the estimation of the joint probability density function of (X,Y). This estimation has to be carried on the known data set. Histogram- and kernel-based pdf estimations are among the most commonly used methods [12]. However, their use is usually restricted to one- or two-dimensional probability density functions (i.e. pdf of one or two variables). However, in the next section, we will use the mutual information for high-dimensional variables (or equivalently for *groups of real valued variables*); in this case, histogram- and kernel-based estimators suffer dramatically from the curse of dimensionality; in other words, the number of samples that are necessary to estimate the pdf grows exponentially with the number of variables. In the context of spectra analysis, where the number of variables grows typically to several hundreds, this condition is not met.

For this reason, the MI estimate used in this paper results from a *k*-nearest neighbour statistics. There exists an extensive literature on such estimators for the entropy [13, 14], but it has been only recently extended to the Mutual Information by Kraskov et al [7]. A nice property of this estimator is that it can be used easily for sets of variables, as necessary in the variable selection procedure described in the next section. To avoid cumbersome notations, we denote a set of real-valued variables as a unique vector-valued variable (denoted X as in the previous section).

In practice, one has at disposal a set of *N* input-output pairs $z^i = (x^i, y^i)$, $i = 1$ to $N$, which are assumed to be i.i.d. (independent and identically distributed) realizations of a random variable $Z = (X, Y)$ (with pdf $\mu_{X,Y}$).

In our context, both X and Y have values in $\Re$ (the set of reals) or in $\Re^p$ and the algorithm will therefore use the natural norm in those spaces, i.e., the Euclidean norm. In some situations (see Section 5.3.1 for an example) prior knowledge could be used to define more relevant norms that can differ between *X* and *Y*. However, to simplify the presentation of the mutual information estimator, we use the same notation for all metrics (i.e., the norm of *u* is denoted $\|u\|$).

Input-output pairs are compared through the maximum norm: if $z = (x, y)$ and $z' = (x', y')$, then

$$\|z - z'\|_\infty = \max(\|x - x'\|, \|y - y'\|). \tag{9}$$

The basic idea of [15, 16, 17] is to estimate *I(X,Y)* from the average distances in the *X*, *Y* and *Z* spaces from $z^i$ to its *k*-nearest neighbours, averaged over all $z^i$.

We now on consider *k* to be a fixed positive integer. Let us denote $z^{k(i)} = \left(x^{k(i)}, y^{k(i)}\right)$ the $k^{\text{th}}$ nearest neighbour of $z^i$ (according to the maximum norm). It should be noted that $x^{k(i)}$ and $y^{k(i)}$ are the input and output parts of $z^{k(i)}$ respectively, and thus not necessarily the $k^{\text{th}}$ nearest neighbour of $x^i$ and $y^i$. We denote $\varepsilon^i = \left\|z^i - z^{k(i)}\right\|_\infty$, $\varepsilon_X^i = \left\|x^i - x^{k(i)}\right\|$ and $\varepsilon_Y^i = \left\|y^i - y^{k(i)}\right\|$. Obviously, $\varepsilon^i = \max\left(\varepsilon_X^i, \varepsilon_Y^i\right)$.

Then, we count the number $n_X^i$ of points $x^j$ whose distance from $x^i$ is strictly less than $\varepsilon^i$, and similarly the number $n_Y^i$ of points $y^j$ whose distance from $y^i$ is strictly less than $\varepsilon^i$.

It has been proven in [7] that *I(X,Y)* can be accurately estimated by:

$$\hat{I}(X, Y) = \psi(k) - \frac{1}{N}\sum_{i=1}^{N}\left[\psi(n_X^i + 1) + \psi(n_Y^i + 1)\right] + \psi(N), \tag{10}$$

where $\psi$ is the digamma function [7] given by:

$$\psi(t) = \frac{\Gamma'(t)}{\Gamma(t)} = \frac{d}{dt}\ln\Gamma(t), \tag{11}$$

with

$$\Gamma(t) = \int_0^\infty u^{t-1}e^{-u}du. \tag{12}$$

The quality of the estimator $\hat{I}(X,Y)$ is linked to the value chosen for *k*. With a small value for *k*, the estimator has a large variance and a small bias, whereas a large value of *k* leads to a small variance and a large bias. As suggested in [18] we have used in Section 5 a mid-range value for *k*, i.e. *k=6*.

## 3. Variable ordering and selection

Section 2 detailed the way to estimate the mutual information between, on one side, an input variable or set of input variables *X*, and on the other side an output (to be predicted) variable *Y*. In the following of this section, estimations of the mutual information will be used to select adequate variables among the initial set of *M* input variables $X = (X_1, X_2, \ldots, X_M)$. As getting exact values of the mutual information is not possible with real data, the following methodology makes use of the estimator presented in the previous section and uses therefore $\hat{I}(X,Y)$ rather than *I(X,Y)*.

As the mutual information can be estimated between any subset of the input variables and the dependant variable, the optimal algorithm would be to compute the mutual information for every possible subset and to use the subset with the highest one. For *M* input variables, $2^M$ subsets should be studied. Obviously, this is not possible when *M* exceeds small values such as 20. Therefore in a spectrometric context where *M* can reach 1000, heuristic algorithms must be used to select candidates among all possible sets. The current section describes the proposed algorithm.

### 3.1. Selection of the first variable

The purpose of the use of mutual information is to select the most relevant variables among the {*X<sub>j</sub>*} set. Relevant means here that the information content of a variable *X<sub>j</sub>* must be as large as possible, what concerns the ability to predict *Y*. As the information content is exactly what is measured by the mutual information, the first variable to be chosen is, quite naturally, the one that maximizes the mutual information with *Y*:

$$X_{s1} = \arg\max_{X_j} \{\hat{I}(X_j, Y)\}, \quad 1 \leq j \leq M \tag{13}$$

In this equation, $X_{s1}$ denotes the first selected variable; subsequent ones will be denoted $X_{s2}, X_{s3}, ...$

### 3.2. Selection of next variables

Once $X_{s1}$ is selected, the purpose now is to select a second variable, among the set of remaining ones $\{X_j, 1 \leq j \leq M, j \neq s1\}$. There are two options that both have their advantages and drawbacks.

The first option is to select the variable (among the remaining ones) that has the largest mutual information with *Y*:

$$X_{s2} = \arg\max_{X_j} \{\hat{I}(X_j, Y)\}, \quad 1 \leq j \leq M, j \neq s1 \tag{14}$$

The next variables are selected in a similar way. In this case, the goal of selecting the most informative variable is met. However, such way of proceeding may lead to the selection of very similar variables (highly collinear ones). If two variables $X_{s1}$ and $X_r$ are very similar, and if $X_{s1}$ was selected at the previous step, $X_r$ will most probably be selected at the current step. But if $X_{s1}$ and $X_r$ are similar, their information contents to predict the dependent variable are similar too; adding $X_r$ to the set of selected variables will therefore not much improve the model. However adding $X_r$ can increase the risks mentioned in Section 1 (colinearity, overfitting, etc.), without necessarily adding much information content to the set of input variables.

The second option is to select in the second step a variable that maximizes the information contained in the *set of selected variables*; in other words, the selection has to take into account the fact that variable $X_{s1}$ has already been selected. The second selected variable $X_{s2}$ will thus be the one that maximizes the mutual information between the set $\{X_{s1}, X_{s2}\}$ and the output variable $Y$:

$$X_{s2} = \arg\max_{X_j}\{\hat{I}(\{X_{s1}, X_j\}, Y)\}, \ 1 \leq j \leq M, j \neq s1 \quad (15)$$

The variable $X_{s2}$ that is selected is the one that adds the largest information content to the already selected set. In the next steps, the $t^{th}$ selected variable $X_{st}$ will be chosen according to:

$$X_{st} = \arg\max_{X_j}\{\hat{I}(\{X_{s1}, X_{s2}, \ldots, X_{s(t-1)}, X_j\}, Y)\}, \ 1 \leq j \leq M, j \notin \{s1, s2, \ldots s(t-1)\} \quad (16)$$

Note that this second option does not have advantages only. In particular, it will clearly not select consecutive variables in the case of spectra, as consecutive variables (close wavelengths) are usually highly correlated. Nevertheless, in some situations, the selection of consecutive variables may be interesting anyway. For example, there are problems where the first or second derivative of spectra (or some part of spectra) is more informative than the spectra values themselves. Local derivatives are easily approached by the model when several consecutive variables are used. Preventing them from being selected would therefore prohibit the model to use information from derivatives. On the contrary, Option 1 allows such selection, and might thus be more effective than Option 2 in such situation. Note that derivatives are given here as an example only. If one knows a priori that taking derivatives of the spectra will improve the results, applying this preprocessing is definitely a better idea than expecting the variable selection method to find this result in an automatic way. Nevertheless, in general, one cannot assume that the ideal preprocessing is known a priori. Giving more flexibility to the variable selection method by allowing consecutive variables to be selected may be considered as a way to automate, to some extend, the choice of the preprocessing.

### 3.3. Backward step

Both options detailed above may be seen as forward steps. Indeed, at each iteration, a new variable is added, without questioning about the relevance of the variables previously selected.

In the second option (and only in this one), such a forward selection, if applied in a strict way, may easily lead to a local maximum of the mutual information. A non-adequate choice of a single variable at any step can indeed influence dramatically all subsequent choices. To alleviate this problem, a backward procedure is used in conjunction with the forward one. Suppose that $t$ variables have been selected after step $t$. The last variable selected is $X_{st}$. The backward step consists is removing one by one all variables except $X_{st}$, and checking if the removal of one of them does increase the mutual information. If several variables meet this condition, the one whose removal increases the more the mutual information is eventually removed. In other words, after forward step $t$:

$$X_{sd} = \arg\max_{X_{sj}}\{\hat{I}(\{X_{s1}, X_{s2}, \ldots, X_{s(j-1)}, X_{s(j+1),\ldots}, X_{st}\}, Y)\}, \quad 1 \leq j \leq t-1. \quad (17)$$

If

$$\hat{I}(\{X_{s1}, X_{s2}, \ldots, X_{s(d-1)}, X_{s(d+1),\ldots}, X_{st}\}, Y) > \hat{I}(\{X_{s1}, X_{s2}, \ldots, X_{st}\}, Y), \quad (18)$$

then $X_{sd}$ is removed from the set of selected variables.

### 3.4. Stopping criterion

Both procedures (Options 1 and 2) must be given a stopping criterion. Concerning the second option, this is quite easily achieved. The purpose being to select the most informative set of variables, i.e. the one having the largest mutual information with *Y*, the procedure is stopped when this mutual information decreases: If after a forward step

$$\hat{I}(\{X_{s1}, X_{s2}, \ldots, X_{st}\}, Y) < \hat{I}(\{X_{s1}, X_{s2}, \ldots, X_{s(t-1)}\}, Y), \tag{19}$$

then the procedure is stopped at step *t*-1.

The first forward procedure is more difficult to stop. Indeed, it may be considered as a ranking algorithm (the variables are ordered according to their mutual information with the output) rather than a selection one. It is suggested to stop the procedure after a number of steps that will be detailed in the next subsection.

### 3.5. Selected set of variables

The two procedures may lead to different sets of variables, namely *A* for Option 1 and *B* for Option 2. As both procedures have their advantages and drawbacks, it is suggested to use the joined $C = A \cup B$ set. In order to keep the computation time within reasonable limits in agreement with the application constraints, it is suggested to keep all variables selected by Option 2 (set *B*), and to add variables from set *A* so that the total number of variables is *P*. The value of *P* is chosen so that $2^P$ runs of a simple algorithm still fit into the simulation time constraints of the application.

The ultimate goal of any variable selection method is to reduce as much as possible the number of variables in order to improve the quality of the model built, and to improve the interpretability of the selected set of variables. The next step is then to build the $2^P$ possible sets of the $P$ variables (by including or not each of them), and to compute the mutual information between each of these sets and the output variable $Y$. The set finally chosen is the one that has the largest mutual information with $Y$.

Trying all $2^P$ possible sets might be seen as a brute-force algorithm. However everything depends on the value of $P$. With $P$ lower than 20, such exhaustive search is possible, and the mutual information of the selected set is obviously higher than the one of sets $A$ and $B$, because the latter are two among the $2^P$ possibilities. There is thus nothing to loose. On the other hand, an exhaustive search on all $2^M$ sets build on the $M$ initial variables is not feasible, since $M$ largely exceeds 20 in spectra. Selection an initial set such as $C$ is thus necessary.

## 4. Nonlinear regression models

Once a reduced set of variables is selected, it is possible to build a nonlinear regression model to predict the dependent (output) variable. When using nonlinear models, one often has to choose their structure, or complexity inside a family of models. For example, when using polynomial models, one has to choose the degree of the polynomial to be used, before learning its parameters. If more complex models are used, as the artificial neural network models that will be detailed in the following of this section, one also has to control the complexity through the number of parameters that will be involved in the model. This concerns the number of hidden nodes in Multi-Layer Perceptrons (MLP) and in Radial-Basis Function Networks (RBFN), the number $k$ of neighbours to be taken into account in a $k$-NN approach, etc.

The structure or complexity of a nonlinear model, as the order of the polynomial, is therefore characterized by a –usually discrete- supplementary parameter. In order to make the difference clear between the traditional parameters of a model, and the one controlling the complexity among a family, we will refer to the latter as *meta-parameter*. Depending on the field, it is also sometimes referred to as *hyper-parameter*, *model structure parameter*, etc.

## 4.1. Learning, validation and test

Choosing an appropriate (optimal) value of the meta-parameter(s) is of high importance. If the complexity of the model is chosen too low, the model will not learn efficiently the data; it will *underfit*, or have a large *bias*. On the contrary, if the model complexity is chosen too high, the model will perfectly fit the learning data, but it will generalize poorly when new examples will be presented. This phenomenon is called *overfitting*, and should of course be avoided, generalization to new data being the ultimate goal of any model.

In order to both learn the parameters of the model, and select on optimal complexity through an adequate choice of the meta-parameter(s), the traditional way to proceed is to divide the available data into three non-overlapping sets, respectively referred to as *learning set* ($L$), *validation set* ($V$), and *test set* ($T$). Once a complexity is defined (i.e. once the meta-parameters are fixed), $L$ is used to set by learning the values of the model parameters. To choose an optimal value of the meta-parameters, one usually has no other choice than comparing the performances of several models (for example several model complexities among a family). Performance evaluation must be done on an independent set; this is the role of V. Of course, for each choice of the meta-parameters, the model parameters must be learned through L.

Finally, T is used to assess the performances of the best model chosen after the validation step. Indeed it is important to mention that the validation step already produces a performance evaluation (one for each choice of the meta-parameters), but that this evaluation is biased with regards to the true expected performances, as a choice has been made precisely according to these performance evaluations. Having an unbiased estimation of the model performances goes through the use of a set that is independent both from L and V; this is the role of T.

More formally, one can define

$$\Omega = \{(x^i, y^i) \in \Re^M \times \Re, 1 \leq i \leq N_\Omega : y^i = f(x^i)\}, \quad (20)$$

where $x^i$ and $y^i$ are the $i^{st}$ M-dimensional input vector and output value respectively, $N_\Omega$ is the number of samples in the set $\Omega$, and $f$ is the process to model. $\Omega$ can represent the learning set L, the validation set V, or the test set T.

The Normalized Mean Square Error on each of these sets is then defined as

$$NMSE_\Omega = \frac{\frac{1}{N_\Omega} \sum_{i=1}^{N_\Omega} (\hat{f}(x^i) - y^i)^2}{\mathrm{var}(y)}, \quad (21)$$

where $\hat{f}$ is the model, and var($y$) is the observed variance of the $y$ output values, estimated on all available samples (as the denominator is used as a normalization coefficient, it must be taken equal in all NMSE definitions; here the best estimator of the variance is used).

Splitting the data into three non-overlapping sets is unsatisfactory. Indeed in many situations there is a limited number of available data; reducing this number to obtain a learning set *L* of lower cardinality means to use only part of the information available for learning. To circumvent this problem, one uses resampling techniques, like *k*-fold cross-validation [19], leave-one-out and bootstrap [20,21]. Their principle is to repeat the whole learning and validation procedure, with different splittings of the original set. All available data are then used at least once for learning one of the models.

For example, in the *l*-fold cross-validation procedure [19], a first part of the data is set aside as test set *T*. Then, the remaining samples are divided into *l* sets of roughly equal sizes. Next, *l* learning procedures are run, each of them taking one of the sets as validation set, and the other *l*-1 ones as learning set. In this way, each sample (except those in the test set) is used both for learning and validation, while keeping the non-overlapping rule. The results ($NMSE_L$ and $NMSE_V$) are the average of the *l* values obtained.

In spectrometry applications, one is usually faced with a very small number of spectra to analyze (some tens or some hundreds, which is very small compared to the high dimensionality of the spectra). For this reason, the averages computed both at the level of equation (21) and in the *l*-fold cross-validation procedure, may be influenced dramatically by a few outliers. Outliers may sometimes be identified a priori (such as spectra resulting from measurement errors, etc.); in other situations however, they cannot be differentiated a priori, but it can be seen in simulations that they influence dramatically the results. For example, outliers may lead to square errors (one term of the sums in the *NMSE*) that are several orders of magnitude larger than the other ones.

Taking averages, in the *NMSE* and in the *l*-fold cross-validation procedure, then becomes meaningless. For this reason it is advisable to eliminate these spectra in an automatic way in the cross-validation procedure. In the simulations detailed in the next section, the samples are ranked in increasing order of the difference (in absolute value) between the errors and their median; the errors are the differences between the expected output values and their estimations by the model. Then, all samples that are above the 99% percentile are discarded. This procedure allows eliminating in an automatic way samples that would influence in a too dramatic way the learning results, without modifying substantially the problem (only 1% of the spectra are eliminated).

In the following, two examples of nonlinear models are illustrated: the Radial-Basis Function Network and the Least-Square Support Vector Machines. These models will be used in the next section describing the experiments.

**4.2. RBFN and LS-SVM**

RBFN and LS-SVM are two nonlinear models sharing common properties, but having differences in the learning of their parameters. The following paragraphs show the equations of the models; details about the learning procedures (how the parameters are optimized) may be found in the references.

The Radial Basis Functions Network (RBFN) is a function approximator based on a weighted sum of Gaussian Kernels [22]. It is defined as:

$$\hat{y}(x) = \sum_{k=1}^{K} \lambda_k \Phi(x, C_k, \sigma_k) + b, \qquad (22)$$

where

$$\Phi(x, C_k, \sigma_k) = e^{-\left(\frac{\|x-C_k\|}{\sqrt{2}\sigma_k}\right)^2}.  \quad (23)$$

As we can see in (22), the model has three types of parameters. The $C_k$ are called centroids; they are often chosen into dense regions of the input space, through a vector quantization technique [23]. The number $K$ of centroids is a meta-parameter of the model. The $\sigma_k$ are the widths of the centroids, i.e. their region of influence. The choice of the $\sigma_k$ is done according to [24]; this last approach has been shown effective but it introduces a new meta-parameter, called Width Scaling Factor (*WSF*), which has to be optimized. The $C_k$ and $\sigma_k$ are chosen regardless of the outputs $y$, based only on the properties of the distribution of the inputs $x$. This makes the fitting of the parameters very fast, compared to other nonlinear models. Finally, the $\lambda_k$ and $b$ are found by linear regression.

An example of RBFN result on a one-dimensional problem is shown in Figure 1.

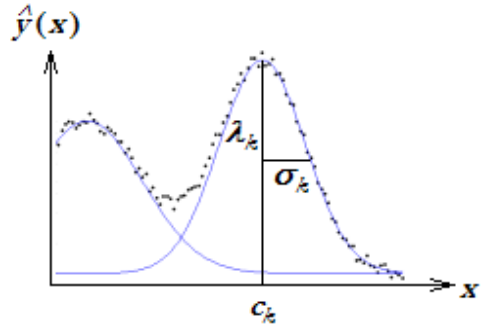

*Figure 1: Example of RBFN. Dots: learning data; solid lines: the two Gaussian functions in the model used to fit the data.*

Both $K$ and the WSF are meta-parameters of the model which are to be optimized through a cross-validation technique. The value of $K$ represents the number of Gaussian Kernels involved in the model. Complex models, with a large number of Kernel functions, have a higher predictive power, but they are more akin to overfitting; their generalization performances are poor. The *WSF* controls the covering of one Kernel function by another. A large value for the *WSF* will lead to a very smooth model, while a smaller value will produce a rather sharp model, with higher predictive performances, but with also more risk to overfit the data.

Least-Squares Support Vector Machines (LS-SVM) [25-27] differ from RBFN in two ways:

- the number of kernels or Gaussian functions is equal to the number *N* of learning samples;
- instead of reducing the number of kernels, LS-SVM introduce a regularization term to avoid overfitting.

The LS-SVM model is defined in its primal weight space by

$$\hat{y}(x) = \sum_{i=1}^{N} \lambda_i \Phi(x, x^i, \sigma) + b, \qquad (24)$$

where $\Phi$ is defined as in (23). The Gaussian kernels are now centred on the learning data $x^i$, and usually a standard deviation $\sigma$ common to all kernels is chosen.

In Least Squares Support Vector Machines for function estimation, the following optimization problem is formulated:

$$\min_{\Lambda, b, e} J(\Lambda, e) = \frac{1}{2} \Lambda^T \Lambda + \gamma \frac{1}{2} \sum_{i=1}^{N} (e^i)^2 \qquad (25)$$

subject to the equality constraints

$$e^i = y^i - \hat{y}(x^i), i = 1, \ldots, N. \qquad (26)$$

Vector Λ is defined as Λ = [$\lambda_1, \lambda_2, \ldots, \lambda_N$], and scalar $\gamma$ is a meta-parameter adjusting the balance between regularization and sum of square errors (first and second parts of equation (25) respectively). Equation (26) must be verified for all $N$ input output pairs ($x^i, y^i$). The set of model parameters consists of vector Λ and scalar $b$.

Solving this optimization problem goes through a dual formulation, similarly to conventional Support Vector Machines (SVM); the solution is detailed in [25-27]. Using the dual formulation allows avoiding the explicit computation of Φ, and makes it possible to use other kernels than Gaussian ones in specific situations. The meta-parameters of the LS-SVM model are the width σ of the Gaussian kernels (taken to be identical for all kernels) and the γ regularization factor. LS-SVM can be viewed as a form of parametric ridge regression in the primal space.

## 5. Experimental results

### 5.1. Datasets

The proposed method for input variable selection is evaluated on two spectrometric datasets coming from the food industry. The first dataset relates to the determination of the fat content of meat samples analysed by near infrared transmittance spectroscopy [28]. The spectra have been recorded on a Tecator Infratec Food and Feed Analyzer working in the wavelength range 850 - 1050 nm. The spectrometer records light transmittance through the meat samples at 100 wavelengths in the specified range. The corresponding 100 spectral variables are absorbance defined by $\log(1/T)$ where $T$ is the measured transmittance. Each sample contains finely chopped pure meat with different moisture, fat and protein contents. Those contents, measured in percent, are determined by analytic chemistry. The dataset contains 172 training spectra and 43 test spectra. A selection of training spectra is given in Figure 2 (spectra are normalized, as explained in Section 5.3.1). Vertical lines correspond to variables selected with the mutual information (MI) method proposed in this paper.

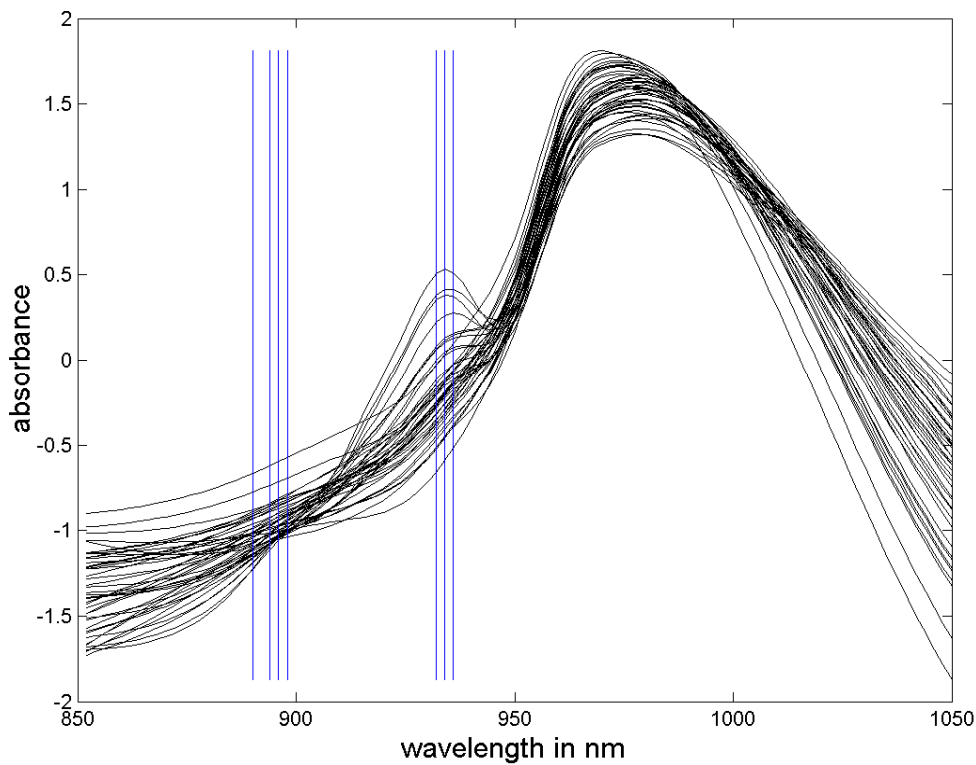

*Figure 2: A selection of spectra from the meat sample dataset (spectra are normalized)*

The second dataset relates to the determination of the sugar (saccharose) concentration in juice samples measured by near infrared reflectance spectrometry [29]. Each spectrum consists in 700 spectral variables that are the absorbance ($\log(1/R)$) at 700 wavelengths between 1100 nm and 2500 nm (where $R$ is the light reflectance on the sample surface). The dataset contains 149 training spectra and 67 test spectra. A selection of training spectra is given in Figure 3. Vertical lines correspond to variables selected with the MI method.

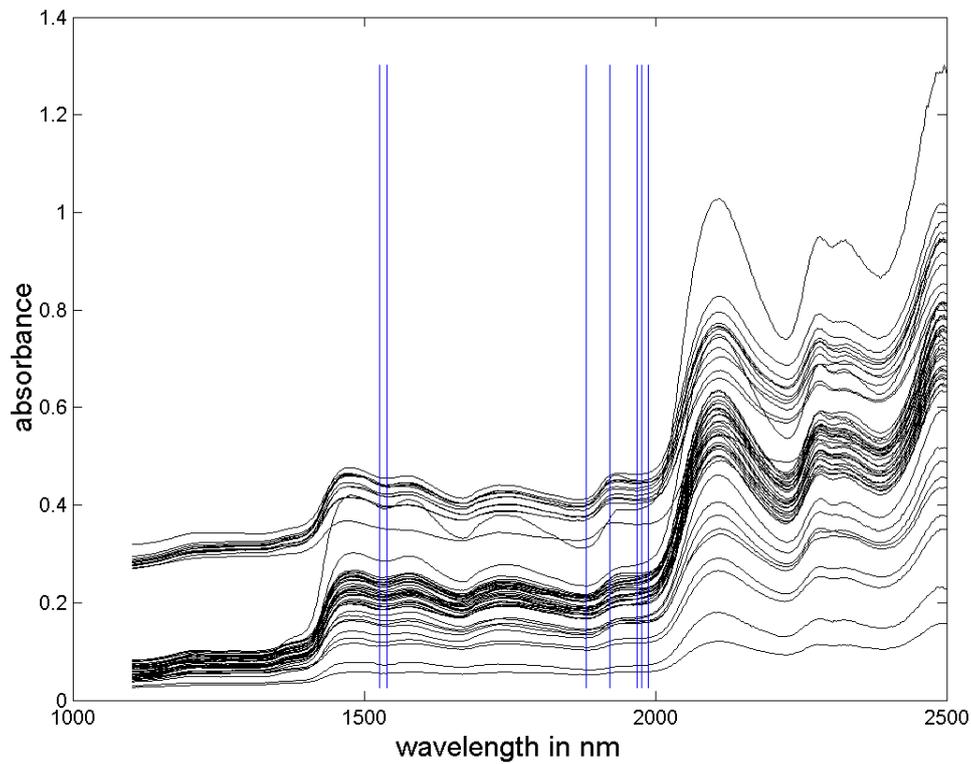

*Figure 3: A selection of training spectra from the juice dataset*

**5.2. Experimental methodology**

The proposed method is benchmarked against several traditional methods that have comparable levels of computation burden. Linear regression methods are also used as reference methods. Each method is designated by a number that will be used in subsequent sections. Figure 4 illustrates the processing steps that compose the compared methods.

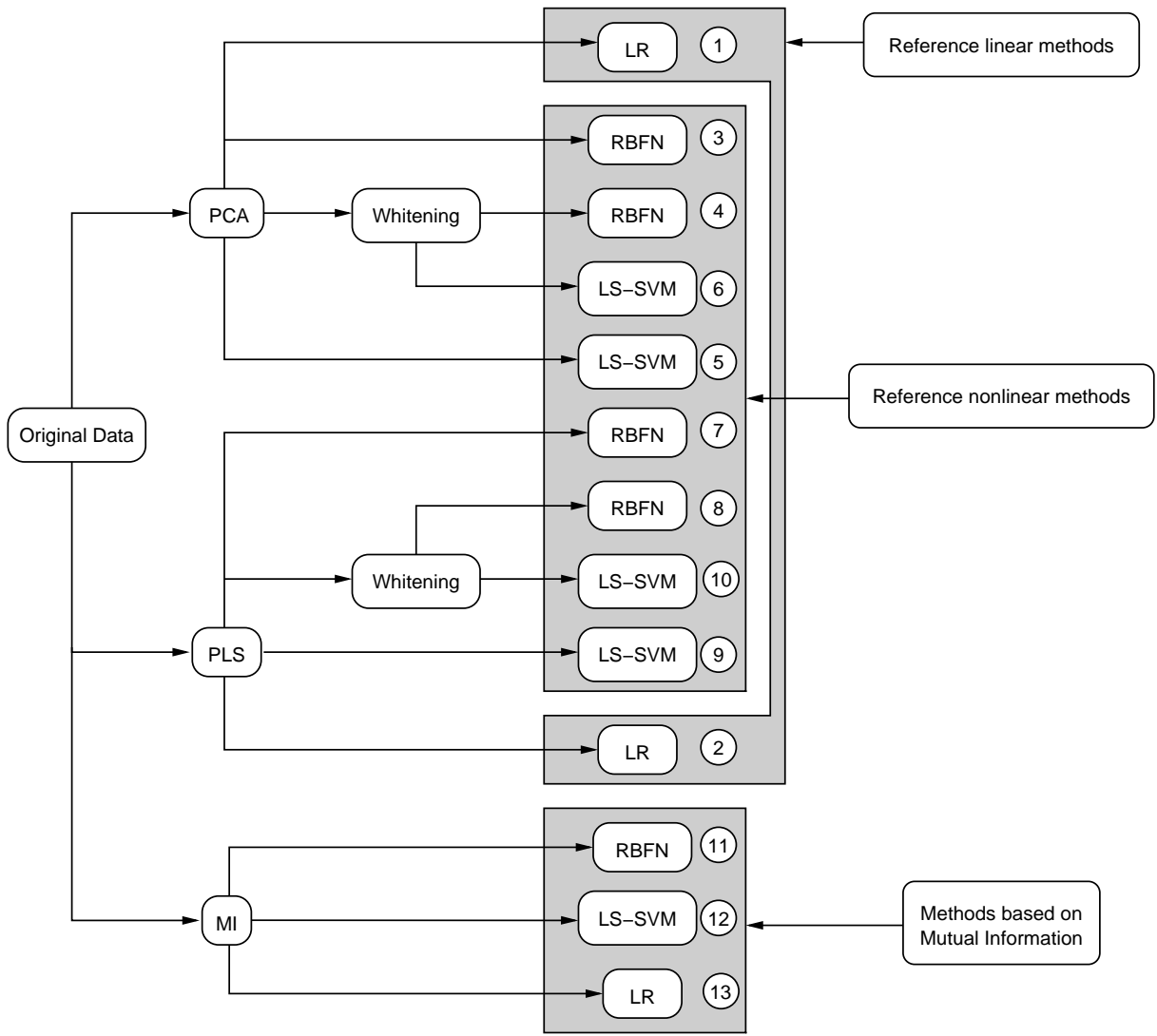

*Figure 4: Summary of data processing methods*

All methods have been implemented in Matlab. The authors of [7,18] provide a mixed Matlab/C implementation [30] of the Mutual Information estimator presented in Section 2.2, restricted to research and education purposes.

**5.2.1. Linear methods**

Reference performances are obtained with two linear regression methods: the Principal Component Regression (PCR) and the Partial Least Square Regression (PLSR). In Method 1, a linear regression model is used to predict the dependant variable (fat content or saccharose concentration) thanks to the first coordinates of the spectra on the principal components produced by a Principal Component Analysis (PCA). The number of principal components is selected by a l-fold cross-validation (as other meta-parameters).

In the second method (Method 2), a standard PLSR is conducted, that is a linear regression model is used to predict the dependant variable thanks to the first factors of the PLS analysis of the spectra. As for the PCR, the number of factors is determined by a l-fold cross-validation.

### 5.2.2 Linear preprocessing for nonlinear models

Obviously, when the problem is highly nonlinear, PCR and PLSR can only provide reference results. It is quite difficult to decide whether bad performances might come from inadequate variables or from the limitation of linear models. Therefore, a fairer benchmark for the proposed method is obtained by using new variables built thanks to PCR and PLSR as inputs to nonlinear models (we therefore use the number of variables selected by l-fold cross-validation). As the same nonlinear models will be used, differences in performances between those methods and the one proposed in this paper will completely depend on the quality of the selected variables. One difficulty of this approach is that nonlinear models such as RBFN and LS-SVM are sensitive to differences in the range of their input variables: if the variance of one variable is very high compared to the variance of another variable, the latter will not be taken into account. Variables produced by PCA or PLS tend to have very different variances as a consequence of their design. Therefore, whitening those variables (reducing them to zero mean and unit variance) might sometimes improve the performances of the nonlinear method.

We thus obtain eight different methods that correspond to all possible combinations of three concepts above: the first one is the linear preprocessing (PCA or PLS), the second one is the optional whitening and the last part is the nonlinear model (RBFN or LS-SVM). Methods are summarized in table 1 (see also Figure 4).

| Method number | Linear method | Whitening | Nonlinear model |
|---|---|---|---|
| (3) | PCA | No | RBFN |
| (4) | PCA | Yes | RBFN |
| (5) | PCA | No | LS-SVM |
| (6) | PCA | Yes | LS-SVM |
| (7) | PLS | No | RBFN |
| (8) | PLS | Yes | RBFN |
| (9) | PLS | No | LS-SVM |
| (10) | PLS | Yes | LS-SVM |

Table 1: linear preprocessing for nonlinear models

### 5.2.3. Variable selection by mutual information

The variable selection method proposed in this paper is used to select relevant inputs for both nonlinear models (RBFN and LS-SVM). As explained in Section 3.5, the last part of the method consists in an exhaustive search for the best subset of variables among $P$ candidates. We choose here to use $P = 16$ variables, which corresponds to 65 536 subsets, a reasonable number for current hardware. As it will be shown in Section 5.3, this number is compatible with the size of set B (Option 2, Section 3): Option 2 selects 8 variables for the Tecator dataset and 7 variables for the Juice dataset.

As explained in Section 2, the mutual information estimation is conducted with $k = 6$ nearest neighbours. The selected variables are used as inputs to the RBFN (Method 11) and to the LS-SVM (Method 12). Again, a l-fold cross-validation procedure is used to tune the meta-parameters of the nonlinear models. We found the LS-SVM rather sensitive to the choice of the meta-parameters and used therefore more computing power for this method, by comparing approximately 300 possible values for the regularization parameter $\gamma$ and 100 values for the standard deviation $\sigma$. The RBFN appeared less sensitive to the meta-parameters. We used therefore only 15 values for the *WSF*. The number $K$ of centroids was between 1 and 30.

As explained in the introduction of this paper, the mutual information models nonlinear relationships between the input variables and the dependant variable. Therefore, the variables selected with the method proposed in this paper should not be suited for building a linear model. In order to verify this point, a linear regression model is constructed on the selected variables (Method 13).

### 5.3. Results

#### 5.3.1. Tecator meat sample dataset

Previous works on this standard dataset (see e.g., [31]) have shown that the fat content of a meat sample is not highly correlated to the mean value of its absorbance spectrum. It appears indeed that the shape of the spectrum is more relevant than its mean value or its standard deviation. Original spectra are therefore preprocessed in order to separate the shape from the mean value. More precisely, each spectrum is reduced to zero mean and to unit variance. To avoid loosing information, the original mean and standard deviation are kept as two additional variables. Inputs are therefore vectors with 102 variables. This method can be interpreted as introduction expert knowledge through a modification of the Euclidean norm in the original input space.

The l-fold cross-validation has been conducted with $l = 4$. This allows to obtain equal size subsets (four subsets with 43 spectra in each) and to avoid a large training time.

The application of Option 2 for the variable selection with mutual information leads to a subset $B$ with 8 variables. Option 1 is then used to rank the variables and to produce another set $A$ such that the union $C = A \cup B$ contains 16 variables. In this experiment, $B$ is a strict subset of $A$, and therefore Option 2 might have been avoided. This situation is a particular case that does not justify using only Option 1 in general. The next section will show that for the juice data set, $B$ is not a subset of $A$. The exhaustive search among all subsets of $C$ selects a subset of 7 variables (see Figure 2) with the highest mutual information.

Among the thirteen experiments described in Section 5.2, four are illustrated by predicted value versus actual value graphs: the best linear model (Figure 5), the best nonlinear model without using the MI selection procedure (Figure 6), and the two models using the MI procedure (Figures 7 and 8).

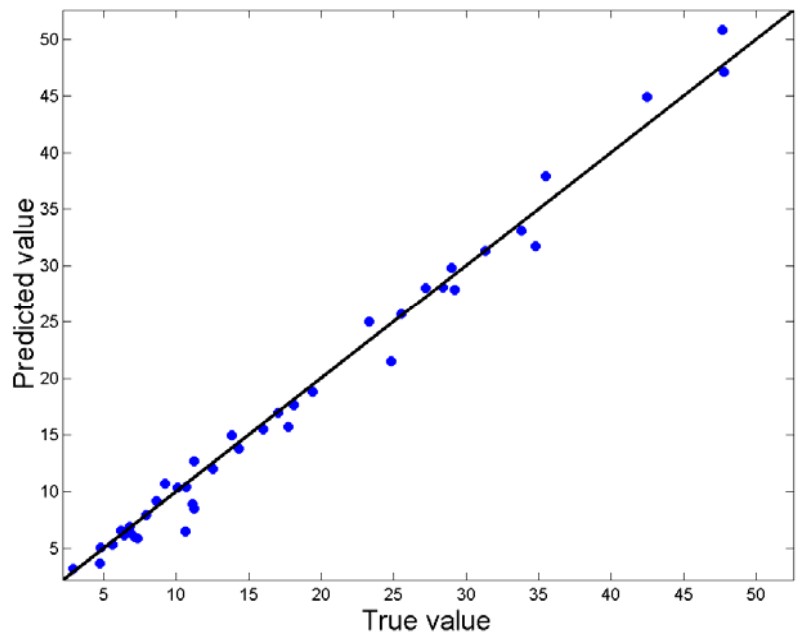

*Figure 5 : Predicted fat content versus the actual fat content with PLSR (Method 2)*

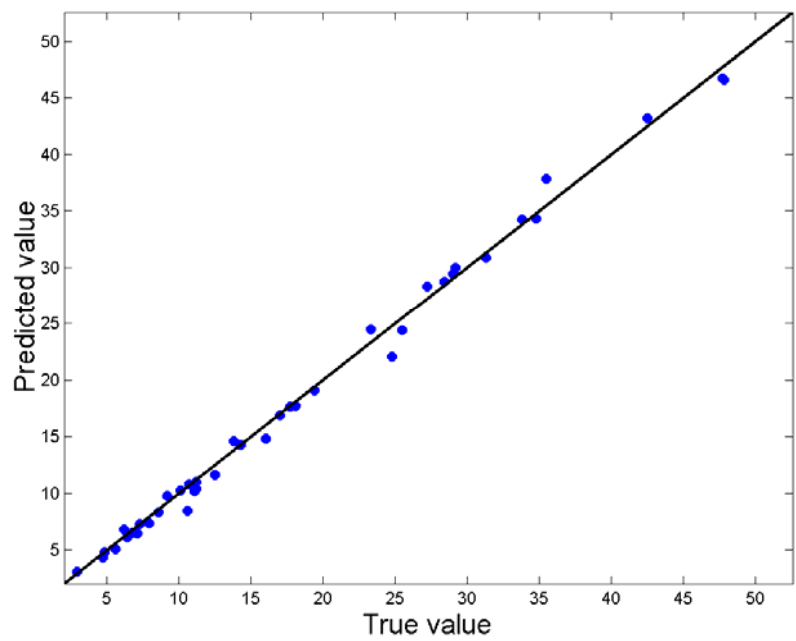

*Figure 6 : Predicted fat content versus the actual fat content with a RBFN on PCA coordinates(Method 3)*

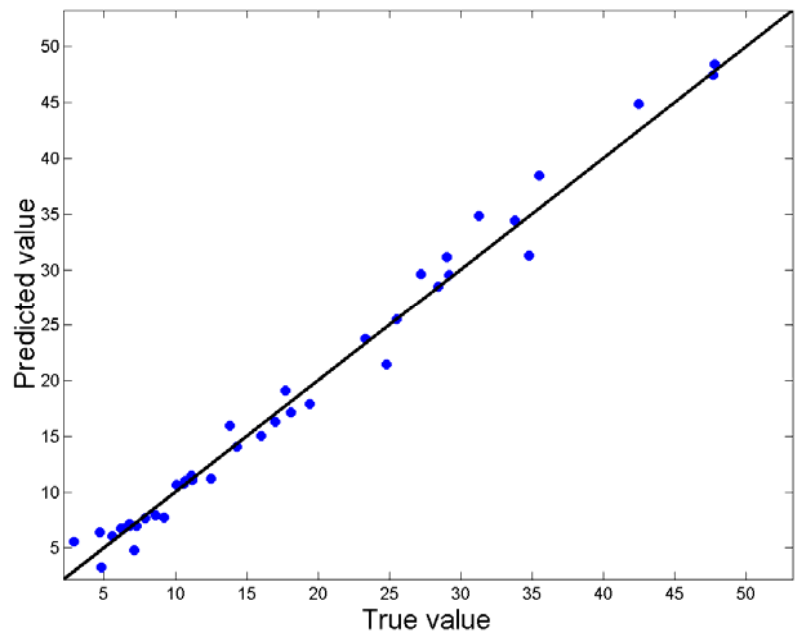

*Figure 7 : Predicted fat content versus the actual fat content with a RBFN on MI selected variables (Method 11)*

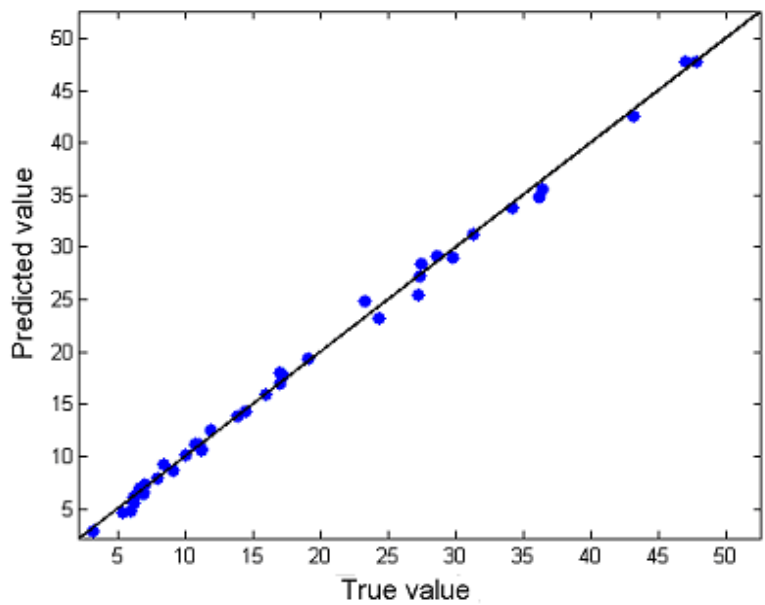

*Figure 8 : Predicted fat content versus the actual fat content with a LS-SVM on MI selected variables (Method 12)*

Table 2 shows the results of the thirteen experiments. All results are given in terms of the Normalized Mean Square Error on the test set. The results in bold correspond to the two best methods.

| Experiment | Preprocessing | Number of variables | Model | $NMSE_T$ |
|---|---|---|---|---|
| (1) | PCA | 42 | Linear | 1.64E-2 |
| (2) | PLS | 20 | Linear | 1.36E-2 |
| **(3)** | **PCA** | **42** | **RBFN** | **4.6E-3** |
| (4) | PCA + whitening | 42 | RBFN | 1.78E-2 |
| (5) | PCA | 42 | LS-SVM | 7.8E-2 |
| (6) | PCA + whitening | 42 | LS-SVM | 6.08E-2 |
| (7) | PLS | 20 | RBFN | 8.27E-3 |
| (8) | PLS + whitening | 20 | RBFN | 5.2E-3 |
| (9) | PLS | 20 | LS-SVM | 1.35E-2 |
| (10) | PLS + whitening | 20 | LS-SVM | 1.12E-1 |
| (11) | MI | 7 | RBFN | 6.60E-3 |
| **(12)** | **MI** | **7** | **LS-SVM** | **2.70E-3** |
| (13) | MI | 7 | Linear | 2.93E-2 |

*Table 2 : Experimental results on the Tecator dataset. See text for details.*

The following conclusions can be drawn.

- The proposed variable selection method gives very satisfactory results, especially for the LS-SVM nonlinear model. For this model indeed, the variable set selected by mutual information allows to build the best prediction. Moreover, the best

- other results obtained by LS-SVM are five times worse than those obtained with the set of variables selected by the proposed method.

- For the RBFN, results are less satisfactory, as the mutual information variable set gives an error that is 1.4 times worse than the one obtained with the principal components selected by PCR. Even so, the final performances remain among the best ones. More importantly, they are obtained with only 7 original variables among 100. Those variables can be interpreted, whereas the 42 principal components are much more difficult to understand.

- Most of the variability in the data is not related to the dependent variable. The PCR has indeed to rely on 42 principal components to achieve its best results.

- The use of PLS allows to extract more informative variables than PCA for a linear model (20 PLS scores give better results than 42 principal components), but this does not apply to a nonlinear model. It is therefore quite unreliable to predict nonlinear model performances with linear tools.

- As the linear models give satisfactory results as illustrated in Figure 4, they also produce correct variables, especially for the RBFN. Even so, the best nonlinear model performs almost six times better than the best linear model. The need of nonlinear models is therefore validated, as well as the need of adapted variable selection methods.

- Method number 13 clearly illustrates the nonlinear aspect of the mutual information. The linear regression model constructed on the variables selected by mutual information has quite bad performances, twice as bad as linear models constructed with PCA and PLS.

### 5.3.2. Juice dataset

For the juice dataset, the actual values of spectral variables appear to be as important as the shape of the spectra. Therefore, no reduction has been implemented and raw spectra are used for all methods. The l-fold cross-validation uses $l = 3$. As for the Tecator dataset, this allows to have almost equal size subsets (two subsets of size 50 and one of size 49) and to avoid an excessive training time.

The same thirteen experiments are carried out on the Juice dataset. Of course, the numbers of variables selected by linear methods differ, but the methodology remains identical. For the MI procedure, the set $B$ of variables obtained with Option 2 contains 7 variables. Option 1 is used to produce a set $A$ such that $C = A \cup B$ contains 16 variables. The exhaustive search among all subsets of $C$ selects a subset of 7 variables (see Figure 2) with the highest mutual information. This final subset contains variables both from $A$ and from $B$: this justifies the use of both options before the exhaustive search.

Table 3 shows the results of those thirteen experiments. All results are given in terms of the Normalized Mean Square Error on the test set. The results in bold correspond to the two best methods.

| Experiment | Preprocessing | Number of variables | Model | $NMSE_T$ |
|---|---|---|---|---|
| (1) | PCA | 23 | Linear | 1.52E-1 |
| (2) | PLS | 13 | Linear | 1.49E-1 |
| (3) | PCA | 23 | RBFN | 1.79E-1 |
| (4) | PCA + whitening | 23 | RBFN | 1.68E-1 |
| (5) | PCA | 23 | LS-SVM | 1.45E-1 |
| (6) | PCA + whitening | 23 | LS-SVM | 1.36E-1 |
| (7) | PLS | 13 | RBFN | 1.54E-1 |
| (8) | PLS + whitening | 13 | RBFN | 2.23E-1 |
| (9) | PLS | 13 | LS-SVM | 1.48E-1 |
| (10) | PLS + whitening | 13 | LS-SVM | 1.51E-1 |
| **(11)** | **MI** | **7** | **RBFN** | **9.86E-2** |
| **(12)** | **MI** | **7** | **LS-SVM** | **8.12E-2** |
| (13) | MI | 7 | Linear | 3.70E-1 |

*Table 3 : Experimental results on the Juice dataset. See text for details.*

The following conclusions can be drawn.

- The best results are clearly obtained with the proposed method, both for LS-SVM and RBFN. The best nonlinear model (a LS-SVM) based on variables selected with the MI method performs more than 1.7 times better than the best nonlinear model (again a LS-SVM) based on principal coordinates.

- The series of experiments shows even more than the Tecator dataset that nonlinear models cannot overcome linear models if they are based on badly chosen variables. Indeed, the best linear model (PCR) obtains almost as good results as the best nonlinear model constructed on linearly chosen variables.

- Even if linear methods allow here a huge reduction in terms of the number of variables (the juice spectra contain 700 spectral variables), the MI algorithm offers better results with fewer variables. Moreover, as those variables are original variables, they can be interpreted much more easily than linear combinations of spectra variables.

- As for the Tecator data set, it clearly appears that variables selected by the MI algorithm are not at all adapted for building a linear model. The results of Method 13 are the worse ones and are more than twice as bad as the results obtained with linear models built with PCA and PLS.

## 6. Conclusions

Selecting relevant variables in spectrometry regression problems is a necessity. One can achieve this goal by projection or selection. On the other hand, problems where the relation between the dependent and independent variables cannot be assumed to be linear require the use of nonlinear models. Projecting variables in a nonlinear way is difficult, a.o. because of the lack of fitness criterion. Moreover, selecting variables rather than projecting them makes it possible to interpret the results, as the wavelength of the selected variables is known.

In this paper we suggest the use of the mutual information to estimate the relevance of spectral variables in a prediction problem. The mutual information has the unique property to be model-independent and able to measure nonlinear dependencies at the same time. We also present a procedure to select the spectral variables according to their mutual information with the output. The procedure is fast, and may be adjusted according to the simulation time constraints of the application.

Finally we show on two traditional benchmarks that the suggested procedure gives improved results compared to traditional ones. The selected variables are shown, giving the possibility to interpret the results from a spectroscopic point of view.

## 7. Acknowledgements

The authors would like to thank Prof. Marc Meurens, Université catholique de Louvain, BNUT unit, for having provided the orange juice dataset used in the experimental section of this paper. The authors also thank the reviewers for their detailed and constructive comments.